\pdfoutput=1
\documentclass[num-refs]{nbdt-article}

\usepackage{hyperref}
\usepackage{siunitx}
\usepackage{times}  %Required
\usepackage{helvet}  %Required
\usepackage{courier}  %Required
\usepackage{url}  %Required
\usepackage{amsmath}
\usepackage{amssymb}

\usepackage{algorithm}
\usepackage[noend]{algpseudocode}

\papertype{Original Article}
\title{Combining imagination and heuristics to learn strategies that generalize}

\author[1]{Erik Peterson PhD}
\author[2]{Necati Alp Müyesser BS}
\author[1,3,4\authfn{1}]{Timothy Verstynen PhD}
\author[1]{Kyle Dunovan PhD}

\affil[1]{Department of Psychology, Carnegie Mellon University, Pittsburgh, PA, 15213, USA}
\affil[2]{Department of Mathematics, Carnegie Mellon University, Pittsburgh, PA, 15213, USA}
\affil[3]{Carnegie Mellon Neuroscience Institute, Carnegie Mellon University, Pittsburgh, PA, 15213, USA}
\affil[4]{Biomedical Engineering, Carnegie Mellon University, Pittsburgh, PA, 15213, USA}

\corraddress{Timothy Verstynen PhD \& Kyle Dunovan PhD, Department of Psychology, Carnegie Mellon University, Pittsburgh, PA, 15213, USA}
\corremail{timothyv@andrew.cmu.edu, dunovank@gmail.com}
\fundinginfo{The research was sponsored by AFOSR/AFRL award FA9550-18-1-0251, NSF CAREER Award 1351748, and a contract with the Army Research Laboratory and accomplished under Cooperative Agreement Number W911NF-10-2-0022.}

\runningauthor{Peterson et al.}

\begin{document}
% \includepdf[pages=1-last]{paper.pdf}
\maketitle

\begin{abstract}
Deep reinforcement learning can match or exceed human performance in stable contexts, but with minor changes to the environment artificial networks, unlike humans, often cannot adapt. Humans rely on a combination of heuristics to simplify computational load and imagination to extend experiential learning to new and more challenging environments. Motivated by theories of the hierarchical organization of the human prefrontal networks, we have developed a model of hierarchical reinforcement learning that combines both heuristics and imagination into a ``stumbler-strategist'' network. We test performance of this network using Wythoff's game, a gridworld environment with a known optimal strategy. We show that a heuristic labeling of each position as hot or cold, combined with imagined play, both accelerates learning and promotes transfer to novel games, while also improving model interpretability.

%\href{http://journals.plos.org/ploscompbiol/article/metrics?id=10.1371/journal.pcbi.1005619}{Standard Structure}

% Please include a maximum of seven keywords
\end{abstract}

\section{Introduction}\label{sec:intro}

\noindent Deep reinforcement learning has shown that it can rival human performance on games of strategy like chess \cite{Silver2017} and Go \cite{Silver2016}, as well as less structured environments like classic Atari games \cite{Zhan2016}. Unlike humans, however, these artificial networks are often unable to transfer performance to new situations, even in environments with relatively trivial changes \cite{Zhang2018,Zhang2018a}. Humans, on the other hand, exhibit transfer without much apparent effort. One part of this innate capacity may come from our ability to imagine new environments and learn from them, a kind of counterfactual reasoning \cite{Pearl2017}. Human success at transferring skills from one context to another may also stem from our ability to develop and exploit simple heuristics \cite{Hart2005,Gigerenzer2014}. Here we combine these two features into a single artificial network, a novel variation of hierarchical reinforcement learning we call the \textit{stumbler-strategist} architecture.

Substantial evidence from cognitive science and neuroscience suggests that strategy learning often relies on a hierarchical form of information processing, with action-value associations learned at lower levels of the hierarchy and abstract predictions about the environment at higher levels \cite{doll12, smittenaar, wunderlich, doll, russek17, odoherty17}. This ability appears to be meditated, at least in part, by the hierarchical organization of prefrontal cortico-basal ganglia-thalamic networks \cite{frank, badre2009, koechlin2007information} that are organized in such a way that anterior regions, reflecting more abstract task representations, have a stronger influence on more posterior regions, reflecting more concrete task goals, than vice versa \cite{verstynen2012}. The prefrontal networks are also thought to have hierarchically organized error signals that facilitate learning complex and conditional rules \cite{alexander2015hierarchical, alexander2018frontal}, and more rostral areas (higher in the hierarchy) appear to be engaged during both rumination \cite{bratman2015nature,kucyi2014enhanced} and imaginative play \cite{carlson2013executive, gonsalves2004neural, schacter2007cognitive}, suggesting a common underlying computation \cite{zarr2019foundations}. Thanks to the phasic dopaminergic signals from midbrain regions \cite{schultz97, eshel2015arithmetic, eshel2016dopamine} these pathways are also capable of learning action values, where rewarded (or punished) actions become more (or less) likely to be executed in the future. This form of learning is analogous to that enacted by deep Q-Learning networks (DQNs) that exhibit behavioral policies determined solely by the feedback of previous actions. 

\begin{figure}[bt]
\includegraphics[width=0.75\linewidth]{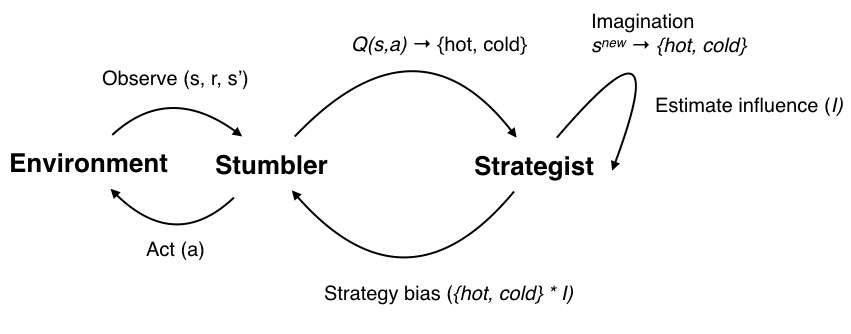}
\centering
\caption{Diagram of the two-layer stumbler-strategist architecture. Stumblers are our name for reinforcement learning agents, driven here by a version of the Q-learning rule and a $\epsilon$-greedy action selection policy. As is typical for such agents the stumblers interact directly with the game environment, learning the state-action values ($Q(s,a)$). The strategist learns only with data extracted from the stumbler. To learn, the strategist first projects $Q(s,a)$ values sampled from the stumbler into one of two classes $\{\text{hot}, \text{cold} \}$ and tries to predict the appearance of these classes in a new, larger, game board. If these predictions are successful, the influence of the strategist is incrementally increased (\textit{Alg. 3}).}
\label{fig:1}
\end{figure}

Taking inspiration from the hierarchical organization of prefrontal cortico-basal ganglia-thalamic networks during learning, we introduce two-layer \textit{stumbler-strategist} networks (Figure \ref{fig:1}), that combine imagination (i.e., internal explorations of environments without directly interacting with the environment) and heuristic learning (i.e., a computationally efficient, but not optimal, solution) in order to find generalizable strategies. Stumblers are reinforcement learning agents that observe and act in the environment. Like any typical reinforcement learning system they only discover the value of specific input-output associations, without appreciation for the organizing features of the game that govern these associations \cite{Sutton2018}. The strategist layer only observes values learned by the stumblers, by sampling, aggregating, and classifying information about the stumblers' actions into one of two classes. They use the heuristic, ``every state is either \textit{cold} or \textit{hot}, good or bad''. The strategist layer never acts directly on the environment but when its confidence is high, it can bias the stumbler's actions. The top-down influence of the strategist is similar to hypothesized role of rostral areas of the human prefrontal cortex \cite{frank2011mechanisms,Badre2012}.

Learning between the layers happens over three stages. First, the stumbler plays the game and collects a set of transitions (state, action, reward) and uses a Q-learning algorithm to learn action-value pairs. Second, the strategist layer takes the current iteration of the stumbler and, from it, samples the expected value for every board position. From the expected value maps it then applies the hot/cold heuristic (see Methods), and learns hot/cold maps by stochastic gradient descent. Finally the current stumbler and strategist layer play a head-to-head game, but on a larger game board than the one either was trained on. As the strategist learns to extrapolate, or transfer, its hot/cold maps to new, never explicitly trained board positions, it will begin to win these head-to-head games. These victories (or losses) are used to update our measure of confidence in the strategist layer, that we term its ``influence''  (Alg.~\ref{algo:strategist-stumbler}.) Importantly, during these sessions where the stumbler and strategist layers play together, learning in the stumbler layer is turned off so that it is not affected by experience in the larger game space during the imaginative play.

Our key innovation is realizing that the simple heuristic used by the strategist-layer lets it accurately extrapolate hot/cold values to new, more challenging, and \textit{never explicitly trained on} board positions. We think of this as a strategic imagination \cite{Weber2017}. The strategist should succeed in extrapolating to new positions because it is asked only to predict a simple binary heuristic, rather than the complex action-value space present in the stumbler (Figure~\ref{fig:5}).

\begin{figure}[bt]
\includegraphics[width=0.5\linewidth]{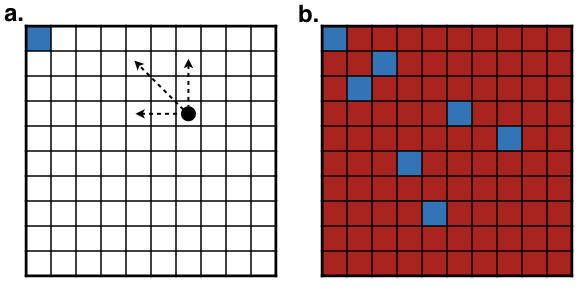}
\centering
\caption{Wythoff's game play. \textbf{a.} Valid moves in Wythoff's are same a Queen's moves in chess. A piece (depicted here as a black circle) can be move laterally, vertically, or diagonally. The goal of the game it to move the piece to the origin of board, shown here as a blue square. \textbf{b,} Optimal play. Optimal moves are to ``cold'' positions (blue). Hot positions are in red. No cold position is accessible from any other cold position, however the winning position is cold. This means the optimal way to play is to always move \textit{to} a cold position forcing your opponent to move \textit{from} a cold position, from which they cannot win.}
\label{fig:2}
\end{figure}

%\subsection{Wythoff's game}\label{subsec:Wythoff}
To isolate strategy learning and transfer, we had our agents play an impartial combinatorial game called Wythoff's game \cite{wythoff1907modification}. Wythoff's game is played on a two dimensional grid in which players alternate turns to move an object that is initially placed randomly on the board. To win, players take turns moving the single game piece to the top-left corner, by moving horizontally, vertically, or diagonally. Moves must be made ``upward'' toward the final winning position, as shown in Figure \ref{fig:2}\textbf{a}. Each player must move when it is their turn. The game ends when one player must take the final position (i.e., top-left position). The player in this position wins the game. Despite their simplicity, these ``gridworld'' environments can provide a challenging but controllable test-bed for transfer learning. In previous work, two state-of-the-art deep reinforcement learning networks could not adapt to subtle differences in the training and testing grid-world environments \cite{Leike2017} suggesting they remain a challenge for deep learning paradigms to overcome.
%are s reasonable paradigm to study transfer given the simplicity of their implementation and specificity as to both what constitutes an ideal strategy and how generalization should be expressed.

Impartial games offer a unique advantage for testing heuristic strategies. The difference between an impartial game, like Wythoff's, and a strategy game like Go, is that an impartial game has a single ground truth solution. Every position in an impartial game is either $\textit{hot}$, meaning that there exists a winning strategy for the player about to make a move, or $\textit{cold}$, meaning that under optimal play, the player about to make a move will always lose. The optimal strategy is always to move from a hot position to cold one. The distribution of $\textit{hot}$ and $\textit{cold}$ positions across the state space in an impartial game usually comes with inherent mathematical structure: $n = m \phi \ \text{,}\ m = n \phi$ where $\phi$ is the golden ratio and $m$ and $n$ are indices on the game board. This structure is visualized in Figure \ref{fig:2}\textbf{b}. Using Wythoff's game, as well as similar impartial games, we explore here how our hierarchical network can learn the heuristic for solving the game and how it can be transferred to different impartial combinatorial games that rely on the same optimal strategy. 

%Our goal here was to see whether our novel strategist-stumbler networks can use heuristics learned from imaginative play (strategist) to help a simple Q-learning network (stumbler) find an optimal strategy that can generalizes across environments. To classify every position on the board, first $Q(s,a)$ values from the stumbler are converted to expected values, where $V(s) = \text{max} Q(s, a)$. $V(s)$ for every $s$ is classified as \textit{bad} if $V(s) < V_\text{bad}$ and \textit{good} if $V(s) > V_\text{good}$, where the thresholds $V_\text{bad}$ and $V_\text{good}$ are hyperparameters of the model. We test this hierarchical network across different board sizes of Wythoff's game show how the heuristic for solving Wythoff's game can be transfered to different impartial combinatorial games that rely on the same optimal strategy. 

% ---------------------------------------------------------------------------------
% ---------------------------------------------------------------------------------
% ---------------------------------------------------------------------------------
\section{Methods}
We have designed a new two-level network architecture that combines traditional $Q$-learning, that we call a stumbler, with a new strategist. The strategist has two key elements: an experimenter provided heuristic, and a input representation that allows the network to imagine, or extrapolate, the value of unseen board positions. These two elements work in combination, and allow our network to directly transfer its knowledge to new, never before experienced, game boards, and rapidly learn new changes to the rules of the game. The heuristic, which projects $Q(s,a)$ values to $\{hot, cold\}$ classes, is described in detail below. In this section we describe the learning algorithms in the stumbler and strategists, as well as the the algorithm we use to control how much influence the strategist has over the stumbler. We conclude this section with a description of our approach to hyper-parameter tuning.

Code for all experiments is available at \url{https://github.com/CoAxLab/azad}

\subsection{Impartial games: Wythoff's game, Nim, and Euclid} \label{sec:impartial_games}
Wythoff's game is played on a two dimensional grid in which players alternate turns to move an object that is initially on the bottom-right corner towards the top-left corner. The player who gets to place the object in the top-left corner terminates, and thereby wins the game. Every turn, the object can be moved horizontally, vertically, or diagonally towards the top-left corner. The distance of each move is only constrained by the requirement that the player cannot move off of the board.

In Wythoff's game the states are all 2-dimensional non-negative integer coordinates. From coordinates $(a, b)$, player 1 ($p_1$) and player 2 ($p_2$) can access all states of the form $(i, b)$, $(a, j)$, and $(a-k,b-k)$ where $0<i<a$, $0<j<b$, and $0<k<min(a,b)$. As mentioned above, every position in an impartial game is either $\textit{hot}$ or \textit{cold}, indicating whether $p_1$ or $p_2$ will win the game under optimal play. The partition of $\textit{hot}$ and \textit{cold} positions in Wythoff's game is deeply embedded in properties of the Fibonacci string and the golden ratio \cite{Aaron2013}. 

The mathematical structure of Wythoff's game manifests in a highly patterned separation of $\textit{hot}$ and $\textit{cold}$ positions. While we benchmark our strategist-stumbler networks on Wythoff's game, we also later impose rule changes to see how well agents trained in Wythoff's game performed in other impartial games that can have a similar \textit{hot}-\textit{cold} structure. 

Nim is an impartial game similar to Wythoff's but where diagonal moves are are disallowed. Like Wythoff's though, and unlike Euclid, as position in a valid direction is allowed. The simple removal of diagonal play profoundly alters optimal play. The optimal way to play a 2d game of Nim is to move such that row and column indices are equal, which amounts to moving to the diagonal positions (via up and left). This is profoundly different than the optimal play in Wythoff's, which centers around position that are multiples of the golden ratio.

Euclid is an extension of Nim where a distance travelled in the horizontal or vertical direction has to be a multiple of the minimum of the horizontal and vertical distance to the top-left corner. Valid positions are those that could be obtained by taking differences between row $m$ and column $n$ when $m$ and $n$ are also multiples of greatest common denominator of $m$ and $n$, (i.e, $(m, n)$ if $\text(GDC(m,n)$.

Note that previous work \cite{Raghu2017} studied Erdos-Selfridge-Spencer games which also have an optimal strategy. The goals of our two papers are different though. Like \textit{Raghu et al (2017)}, we focus on studying transfer compared to an optimal player,  move-by-move. Unlike this previous study, we try and transfer learning to new environments, rather than focusing only on transferring between changes in opponent strategy. 

\subsection{A hot and cold heuristic}
The explicit and very human heuristic used by the strategist is, ``all positions are either good or bad''. To fit the game good positions are \textit{hot} and bad positions are \textit{cold}. To classify every position on the board, first $Q(s,a)$ values from the stumbler are converted to expected values, where $V(s) = \text{max} Q(s, a)$. A $V(s)$ is classified as \textit{bad} if $V(s) < V_\text{bad}$ and \textit{good} if $V(s) > V_\text{good}$, where the thresholds $V_\text{bad}$ and $V_\text{good}$ are hyperparameters of the model.

\subsection{Network design}\label{sec:net_design}
Stumbler learning is governed by Q-Learning, extended to allow for ``top-down'' strategist feedback and for mutual action observations, where learning updates happen over player-opponent joint-action pairs. Here the stumbler was implemented as a look-up table, defined by the Algorithm \ref{algo:stumbler}.

\begin{algorithm}
\caption{Learning algorithm used by stumbler}
\begin{algorithmic}[1]
\Procedure{Stumble}{$Q$}
    \State $n \gets$ Learning episode
    \State $\epsilon_0 \gets$ Max exploration-exploitation 
    \State $\gamma \gets$ Value bias
    \State $\alpha_s \gets$ Stumbler learning rate
    \State $I \gets$ Strategist influence
    \State $B \gets$ Strategy bias, given $\text{Strategist}$
    \State $G \gets$ Initialize Wythoff's game 
    \State $s \gets$ A position in $G$
    \While{$G$ continues} 
        \State $\epsilon \gets$ $\frac{\epsilon_0}{\text{log} n + e}$ \Comment{Anneal}
        \State $action \gets$ $\epsilon\text{-greedy}(Q(s,action) + I * B(s), \epsilon)$ 
        \State \textbf{do} $action$ on $G$, update $s$
        \If{$G$ ends} 
            \State $reward \gets 1$ \Comment{Winning move}
        \Else \Comment{Opponent plays}
            \State \textbf{do} $action$ on $G$, update $s$
            \If{$G$ ends} 
            \State $reward \gets -1$ \Comment{Opponent wins}
            \Else 
            \State $reward \gets 0$
            \EndIf
        \EndIf 
        \State $Q' \gets$ max Q-value from $s'$ \Comment{Joint-action update} 
        \State $Q(s,action) \gets \alpha (reward + \gamma Q' - Q(s, action))$ 
        \State $I \gets \text{Influence}(\text{Stumbler}, \text{Strategist})$
    \EndWhile
    \State \textbf{return} $Q$
    \EndProcedure
\label{algo:stumbler}
\end{algorithmic}
\end{algorithm}

%\subsection{Strategist layer learning}
The strategist is a two-layer neural network, trained on input coordinates $(i,j)$ and output values derived using the \textit{hot}-\textit{cold} heuristic described in Section \ref{sec:impartial_games}. Its behavior is governed by Algorithm \ref{algo:strategist}. Learning in the full network relied on error backpropagation with stochastic gradient descent, and a learning rate of $\alpha_r$. Training set batch sizes were half the size of the dataset and were sampled with replacement.

\begin{algorithm}[bt]
\caption{Learning algorithm used by the Strategist}
\begin{algorithmic}[1]
\Procedure{Strategist}{$Stumbler$}
    \State $\alpha_r \gets$ Strategist learning rate
    \State $n_r \gets$ Number of training episodes
    \State $n \gets$ Episode counter
    \State $B \gets$ Strategist bias
    \State $Q \gets$ Complete $Q$-table from $Stumbler$
    \State $dataset \gets \text{HotColdHeuristic(Q)}$ \Comment{See main text}
    \State $model \gets$ initialized neural network with default settings
    \While{$n < n_r$} 
        \State $trainset \gets$ \textbf{\textit{sample}}$(dataset)$
        \State \textbf{\textit{backpropagate}} $model$ with $trainset$ through cost func
    \EndWhile
    \State $B \gets$ $model$ for all $s$
    \State \textbf{return} $\text{B}$
\EndProcedure
\label{algo:strategist}
\end{algorithmic}
\end{algorithm}

%\subsection{Strategist layer influence}
To judge how much influence the strategist should have over the the stumbler, the two layers play a single game of Wythoff's using a purely greedy strategy (see Algorithm \ref{algo:strategist-stumbler}). They play on a game board larger than the one the stumbler was trained on. The intuition behind this approach is that if the strategist has useful transferable knowledge it should soundly defeat the stumbler on this larger and new game. Every time the strategist wins, it's influence, $I$, over the stumbler increases by $\alpha_I$. Every time the strategist looses its influence declines by the same amount.

\begin{algorithm}[bt]
\begin{algorithmic}[1]
\Procedure{Influence}{$Stumbler, Strategist$}
    \caption{Strategist influence algorithm}\label{influence}
    \State $\alpha_I \gets$ Influence learning rate
    \State $I \gets$ Influence 
    \State $win \gets$ Strategist score
    \State $G \gets$ Initialize Wythoff's game 
    \State $s \gets$ A position in $G$
        \While{$G$ continues} 
            \State $action \gets$ $\text{greedy}(\text{Stumbler}(s))$
            \State \textbf{do} $action$ on $G$, update $s$
            \If{$G$ ends} 
                \State $win \gets 0$
            \EndIf 
            \State $action \gets$ $\text{greedy}(\text{Strategist}(s))$
            \State \textbf{do} $action$ on $G$, update $s$
            \If{$G$ ends} 
                \State $win \gets 1$
            \EndIf 
        \EndWhile
    \If{$win > 0$} 
        \State$I \gets I + \alpha_I$
    \Else 
        \State$I \gets I - \alpha_I$
    \EndIf 
    \State $I \gets \text{clip}(I,-1,1)$ \Comment{I is limited}
    \State \textbf{return} $I$
\EndProcedure
\label{algo:strategist-stumbler}
\end{algorithmic}
\end{algorithm}

The strategist does not have access to training data besides that provided by the stumbler layer. The strategist learns to play only from the stumbler. The simple nature of what it learns (hot/cold) though lets it correctly infer play on larger boards, as we'll show. Optimal play in Wythoff's is based on golden ratio positions for any board size. The strategist learns this invariant strategy, learning in effect the golden ratio. We believe that learning this invariance requires we simplify the learning problem with a compatible heuristic.

In some runs (Figure~\ref{fig:5}) we replace the learned strategist with a perfect strategist with ideal \textit{hot}/\textit{cold} values hard-coded into the network. This fixed optimal network served as an ideal oracle or reference to judge good learning performance. Hard-coding in \textit{hot}/\textit{cold} values, shows us the maximum rate at which the strategist can improve performance of the stumbler. The closer a model is to this, the better.

\subsection{Network training and parameter tuning}
All networks were trained on a fixed number episodes, which preliminary runs showed were well past the learning plateau. The exception to this were DQN control experiments described in Figure~\ref{fig:9}. Stumbler-only training used this iteration count directly. The stumbler-strategist network required a nesting of the training procedure, which worked as follows. To begin training first the stumbler would play $n_s$ games and learn from the each game independently (Algorithm \ref{algo:stumbler}). Then the strategist takes in the stumbler, extracts its value estimates, applies the heuristic (see \textit{A hot and cold heuristic} above) generating a dataset, and would train on $n_r$ samples of this dataset (Algorithm \ref{algo:strategist}). Finally the influence of the strategist was estimated and applied downstream (Algorithm \ref{algo:strategist-stumbler}). This overall pattern then repeated for $n$ iterations, but the total number of training iterations $n * n_r * n_r$ was constrained to be 75,000 episodes. 

%\subsection{Parameter tuning}
Overall, network performance was robust to a wide range of hyperparameter settings. Parameter tuning was done via grid search, carried out piece-wise. First, the stumbler was tuned. Second, the strategist's learning and heuristic parameters were tuned. Next $V_{hot}$ and $V_{cold}$ were optimized followed by the influence rate ($\alpha_I$). Finally, the depth and unit number of the strategist was optimized. Each stage relied, in part, on the previous tuning stages. The stumbler was implemented as simple one-hot look-up table, and so had no internal parameters. The optimal hyperparameter configuration we arrives at is found in \textit{Table 1}.

\begin{table}[h]
\caption{Network hyperparameters}
\label{table:1}
\begin{tabular}{lll}
\textbf{Meaning} & \textbf{Symbol} & \textbf{Value}                          \\
Stumbler learning rate                                  & $\alpha_s$ & 0.4   \\
Strategist learning rate                                & $\alpha_r$ & 0.025 \\
Influence learning rate                                 & $\alpha_I$ & 0.2   \\
Exploration-exploitation                                & $\epsilon$ & 0.4   \\
Value bias                                              & $\gamma$   & 0.5   \\
Stumbler iterations (strategist only)                   & $n_s$      & 500   \\
Strategy iterations (strategist only)                   & $n_r$      & 500   \\
Hot threshold                                           & $V_{hot}$ & 0.5   \\
Cold threshold                                          & $V_{cold}$  & -0.5  \\
Influence (initial)                                     & $I$        & 0.0   \\
Strategist bias (initial)                               & $B$        & 0.0  
\end{tabular}
\end{table}

\begin{figure}[p!]
\includegraphics[width=0.7\linewidth]{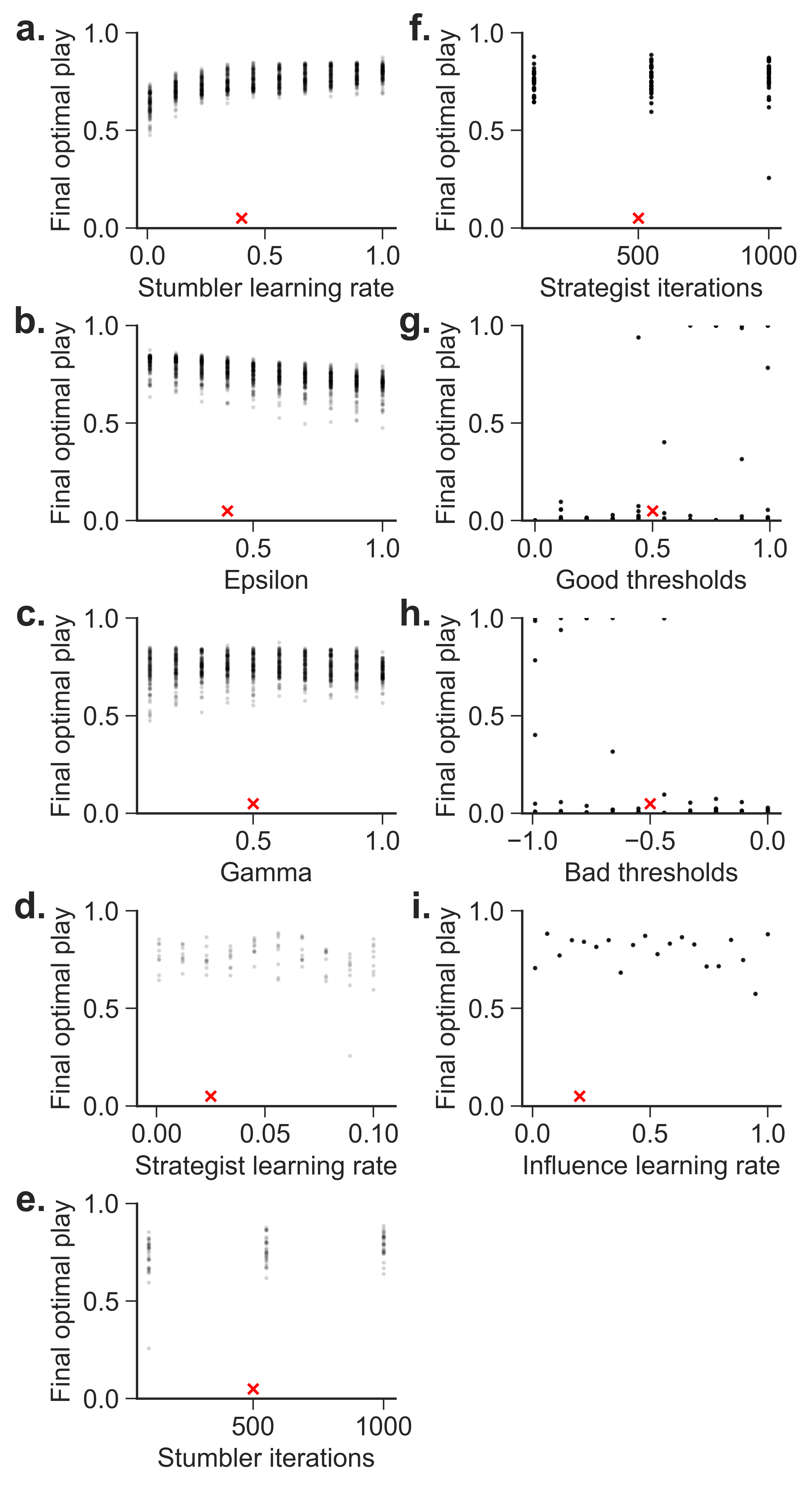} 
\centering
\caption{
Hyperparameter tuning.
\textbf{a.} Stumbler learning rate
\textbf{b.} Epsilon
\textbf{c.} Gamma
\textbf{d.} Strategist learning rate
\textbf{e.} Stumbler iterations
\textbf{f.} Strategy iterations
\textbf{g.} Hot threshold
\textbf{h.} Cold threshold
\textbf{i.} Influence learning rate
The red ``x'' in all figures is the optimal or reference parameter value used in all models, unless noted otherwise. The search strategy for these values is described in \textit{Parameter tuning}. Reference values are in \textit{Table 1}.
}
\label{fig:8}
\end{figure}

%\subsubsection{Stumbler tuning}
In tuning the stumbler we explored the parameter for the greediness of the selection policy, $\epsilon$, from $(0.1-1)$, $\alpha_s$ from $(0.01,1)$, and $\gamma$ from $(0.1-1)$. In taking 10 samples from each range and searching the full permutation space, we sampled a 1000 hyperparameter combinations in tuning the stumbler. The stumbler-strategist tuning stage one explored learning rate $\alpha_r$ and the training iteration numbers $n_s$ and $n_r$, over the following respective ranges, (0.001-0.1), (100,1000), and (100,1000) forming a 10x3x3 sampling space (Figure \ref{fig:8}\textbf{b}). Stage two searched $V_{cold}$ and $V_{hot}$ from $(-1,0)$ and $(0,1)$ forming a 10x10 sampling space. Stage three tuned the influence rate $\alpha_I$ $(0.01-1.0)$ over 20 samples. The final stage explored the depth and width of the two-layer strategist from $n_{\text{hidden1}}$ from $(15, 500)$ and $n_{\text{hidden2}}$ from $(0, 50)$, in 10x10 sampling space (not shown). As is clear in Figure \ref{fig:8}, there is substantial degree of slackness or robustness in parameter choices. As such we hand picked middle values from each ``hot'' region (Table 1). For more information on the meaning of these parameters see Section \ref{sec:net_design}. Note that temporal annealing of $\epsilon$ (Algorithm \ref{algo:stumbler}) is \textit{required} for convergence to optimal play.

\subsection{Deep Q-Learning Network (DQN)}
We compared the performance of our stumbler-strategist network against several DQN \cite{Mnih2013,minvanHasselt2015} baselines. In designing these networks we searched both architectures and hyperparameters to identify a single model who performed best (see Table~\ref{table:2}). We then tested this network's transfer performance measuring both optimal play and play against a random opponent.

% Finding optimal play in Wythoff's requires seeing the whole game board, while standard convolutional networks, by design, identify local invariants. A multi-layer perceptron (MLP), by contrast, can learn using a global view.

The network architectures fell into one of three classes -- \textit{xy}, \textit{hot}, and \textit{conv}. The first \textit{xy} was a multilayer perceptron (MLP) whose input is a vector of game position and action, represented as a set of Cartesian coordinates. Its output was a scalar, representing a Q-value estimate. This was akin to our strategist's representation. The second \textit{hot} was also an MLP but it used a one-hot representation. Its input was a $N x N$ vector, that was zeros except for a 1 placed in the current board position. The output was a $N x N$ vector, representing the value of all possible moves at that position. (Illegal moves were masked.) The third \textit{conv} had the same input/output representation as \textit{hot}, but featured convolutional layers whose ability to learn local features could, in principle, allow the network to generalize to new board sizes and rules. The convolutional networks featured some number of hidden convolutional layers, followed by a dense layer, and a linear readout head. The hidden layers we considered for all these three classes are described in Table~\ref{table:2}.

In tuning DQ networks we explored $\epsilon$ from $((0.1, 0.5)$, $\alpha_s$ from $(0.0025, 0.25)$, and $gamma$ from $(0.1, 0.5)$. We first conducted a grid search taking between 5 and 50 samples from each range, and searching its full permutation space. We sampled a about 1250 hyperparameter combinations in tuning each DQN. Hyperparameter runs were independently for the architectures. Like in the tabular stumbler, the exploration term $\epsilon$ was annealed during training. Following the grid search, we choose the two most promising models based on optimal play score (\textit{xy1} and \textit{xy4}) and used these as the basis for population-based training, using the Optuna library \cite{Akiba2019}. The final best model, according to its average optimal play score, we named \textit{optuna} (Table.\ref{table:2}).

\begin{table}[h]
\caption{DQ network architectures}
\label{table:2}
\begin{tabular}{llll}
Name   & Type & Hidden layers          & Features (layer1,layer2,..) \\
xy1    & MLP  & 1                      & 15                          \\
xy2    & MLP  & 1                      & 100                         \\
xy3    & MLP  & 2                      & 10,20                       \\
xy4    & MLP  & 3                      & 100,25,25                   \\
xy5    & MLP  & 2                      & 1000,2000                   \\
optuna & MLP  & 3                      & 10,11,13                    \\
hot1   & MLP  & 1                      & 15                          \\
hot2   & MLP  & 1                      & 100                         \\
hot3   & MLP  & 2                      & 10,20                       \\
hot4   & MLP  & 3                      & 100,25,25                   \\
hot5   & MLP  & 2                      & 1000,2000                   \\
conv1  & Conv & 5 (conv: 1-3, mlp 4-5) & 8,16,16,5184,20             \\
conv2  & Conv & 5 (conv: 1-3, mlp 4-5) & 32,64,64,5184,512          
\end{tabular}
\end{table}

\section{Results}
By design (Section \ref{sec:net_design}), the strategist never directly observes or acts on the game during learning. Instead it tries to extrapolate, or imagine, the $\{hot, cold\}$ values on a larger game board that the stumbler never encounters. This extrapolation is possible because the complex $Q(s,a)$ value structure is reduced to a binary $\{hot, cold\}$ representation. This, in turn, is based on the heuristic that each position can be exclusively either hot or cold. Our hope was that our heuristic would naturally map onto the \textit{hot}/\textit{cold} board structure of Wythoff's game (See the Section \ref{sec:impartial_games} above for more).

\begin{figure}[bt]
\includegraphics[width=0.6\linewidth]{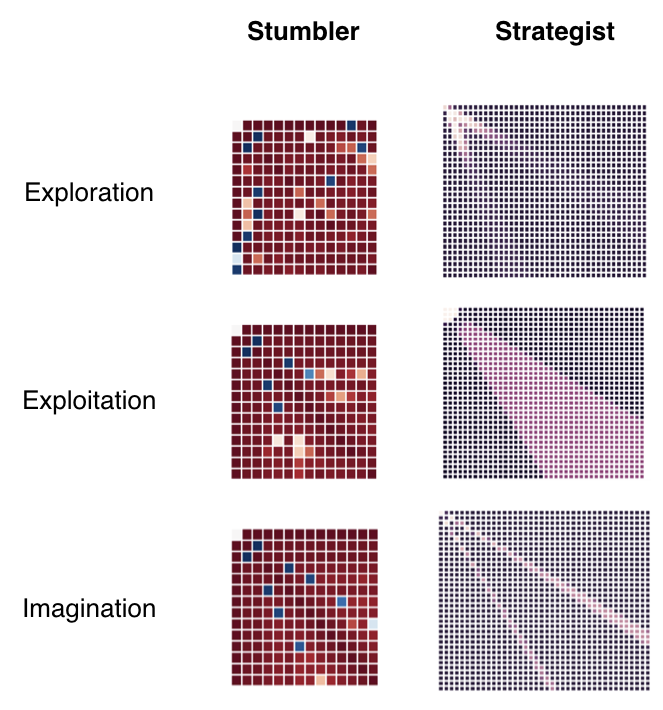}
\centering
\caption{Learning in three stages of training on Wythoff's game. The early models (exploration) will be largely unsuccessful, while certain inaccurate transfers (exploitation) will supply reasonable strategies to the stumbler, allowing the provision of useful datasets into the network that translate into accurate and general models (imagination). In this example the stumbler trained on a 14 by 14 board for 2000 game-plays, across each strategist time-step. The strategist learned to play on a 50 by 50 game board.
}
\label{fig:3}
\end{figure}

Figure \ref{fig:3} shows how the two layers of the network value board positions at different stages of learning. Models that are developed in the earlier stages of training remain mostly irrelevant to transfer, meaning that the estimation of $\{hot, cold\}$ spaces remains largely local to the area of the board that the stumbler interacts with. Soon after initial training, models begin to meaningfully generalize, although still with low accuracy. Such models are crucial for the learning process because they influence the way that the stumbler chooses to explore different action spaces. Without such guidance, the stumbler explores actions without any overall purpose or insight (hence our label for this layer). Yet, with the guidance from the strategist, the stumbler explores actions that would either contradict or confirm the strategist's imagined hypothesis about the nature of the learning environment that it cannot directly learn from. This results in a network that eventually converges on the optimal strategy for identifying $\{hot, cold\}$ spaces.

\begin{figure}[bt]
\includegraphics[width=0.5\linewidth]{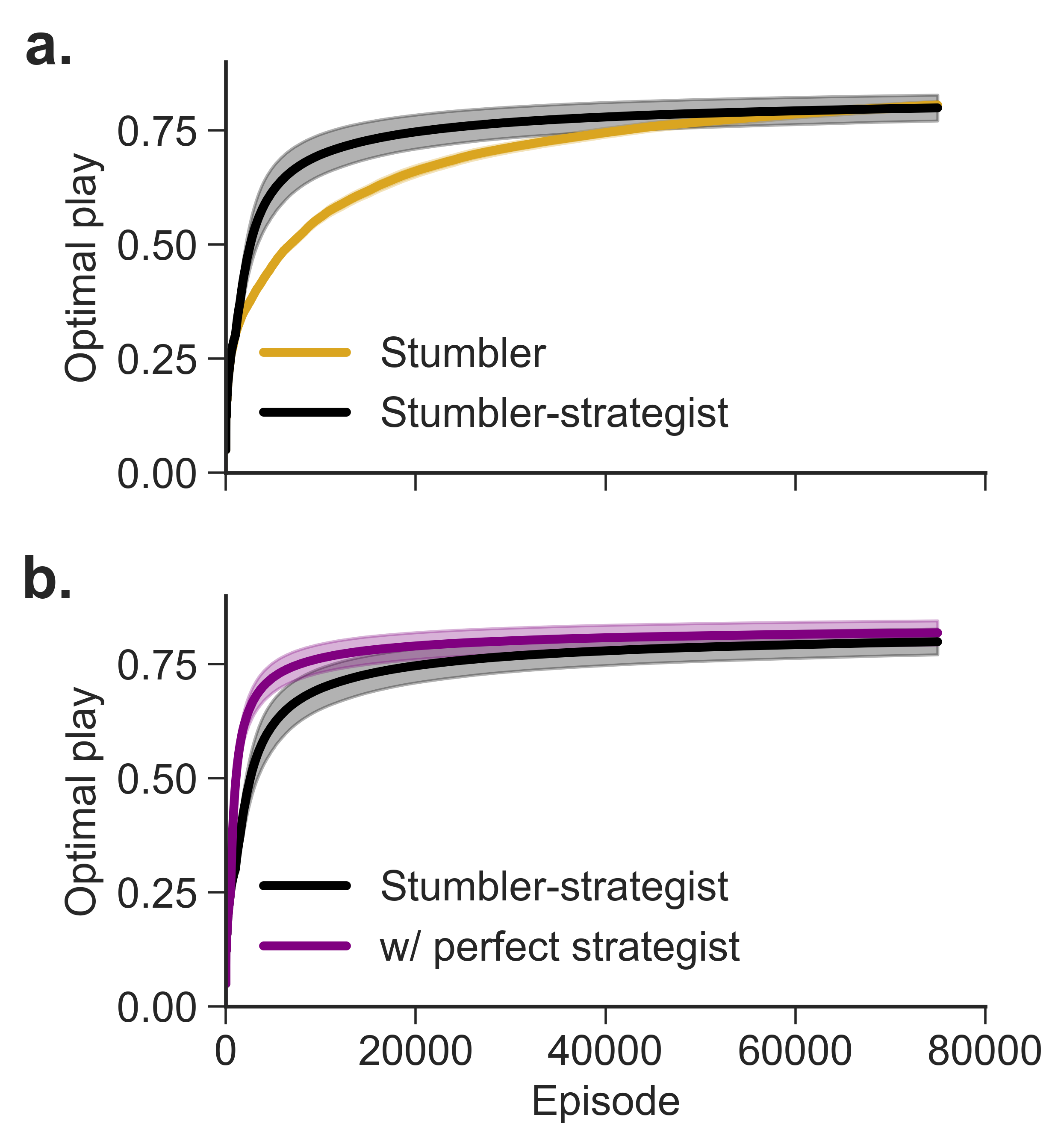}
\centering
\caption{\textbf{a.} Learning optimal moves in Wythoff's game with (black) and without (gold) strategist. \textbf{b.} Learning when a stumbler is biased by a perfect strategist's (purple), compared to the learned strategist from \textit{a.} (black). Error bars denote 2 * standard error, and were calculated from 20 unique random seeds.
}
\label{fig:4}
\end{figure}

%\subsection{Stumbler-strategist performance}
In order to evaluate how much of this performance actually depends on the strategist, we compared the performance of the stumbler alone to the stumbler-strategist network. We only evaluated performance within regions where the stumbler has received feedback. Figure~\ref{fig:4}\textbf{a} shows the fraction of moves that were optimal for both networks during learning. Initial learning in the strategist-stumbler network is more than twice as fast as the stumbler alone. Both networks eventually plateau at the same level of final performance, indicated by the overlapping curves over the last few thousand games in Figure~\ref{fig:4}\textbf{a}. Thus having the strategist accelerates learning of the stumbler within the confines of the original training space.

We next set out to see how well the stumbler-strategist performs against the best possible agent. For this we compared learning between the trained strategist-stumbler and a perfect oracle. Our perfect oracle agent here is a variant of the strategist-stumbler where the optimal strategy for playing Wythoff's is hard coded into into the strategist.  Figure~\ref{fig:4}\textbf{b} shows that the trained strategist-stumbler shows only a slightly lower performance during the early training episodes when compared against the perfect oracle agent. Performance quickly converges between the two agents, suggesting near-optimal accuracy in the trained full network. 

\begin{figure}[bt]
\includegraphics[width=.95\linewidth]{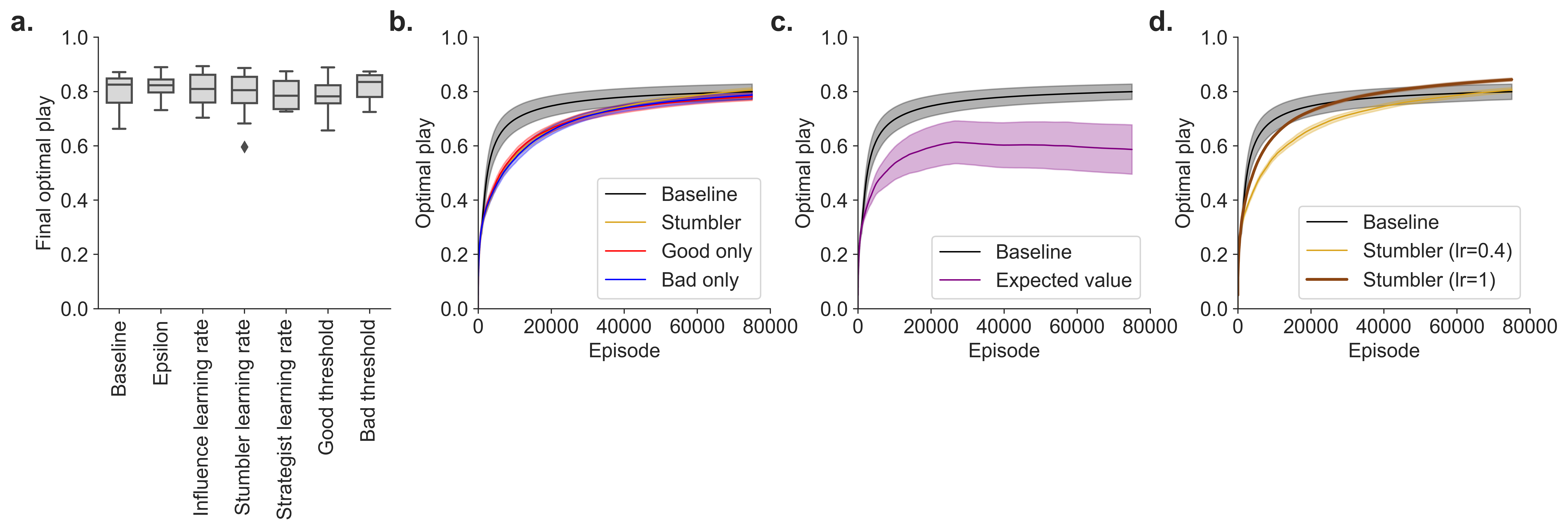}
\centering
\caption{Control experiments. \textbf{a.} Hyper-parameter sensitivity. \textit{Baseline} is 20 standard runs, randomly initialized, using optimal hyperparameters (whose selection is describe in the Methods section). These optimal values are found in Table 1. Epsilon ($\epsilon$) is the exploration parameter in the standard $\epsilon$-greedy algorithm (sampled here from $0.01-0.8$). \textit{Influence learning rate} ($\alpha_I$) controls quickly the strategists bias effects the stumblers actions ($0.01-0.4$). \textit{Stumbler learning rate} is the stumblers learning rate ($\alpha_s$), which in typical Q-learning is often denoted by $\alpha$ ($0.2-0.6$). The \textit{strategist learning rate}  ($\alpha_r$) controls in the the deep strategist ($0.01-0.05$; values larger than 0.08 lead to catastrophic failure). The \textit{hot} and \textit{cold threshold} are the value thresholds, $V_\text{hot}$ and $V_\text{cold}$ (sampled from $0.0-0.5$ and $-0.5-0$ respectively). 
\textbf{b.} Effect of alternate heuristic choices. Our standard heuristic maps $Q(s,a)$ values to \textit{hot}/\textit{cold} classes. Here we explored an alternate approach where the strategist predicts only \textit{hot} or only \textit{cold}, rather than both. 
\textbf{c.} Effect of having no heuristic. In this model the strategist learned the expected value of positions, rather than hot/cold classes. 
\textbf{d.} Effect of having a higher learning rates in a stumbler-only model. It could be the increases in learning performance we observed in the previous figure were not due to hot/cold projection in the strategist layer, but came indirectly with the strategist acting to simply increase the effective learning rate of the stumbler. We ruled this out by choosing the largest rate possible (lr=1) for stumbler-only learning.
Error bars denote 2 * standard error.}
\label{fig:5}
\end{figure}

%\subsection{Parameter sensitivity and heuristic design}
To ensure that the observed performance of the stumbler-strategist network was not specific to a particular set of parameters, we evaluated final optimal play performance over the last 100 episodes as parameters of the network were independently perturbed. The performance of these perturbed networks was compared against a baseline run using standard training parameters (\textit{Table 1}.), but run with 20 different random seeds while parameter perturbations used a fixed seed. Intuitively, if the performance variability due to parameter changes is similar to random behavioral changes, then we can consider that parameter choice to be robust. Indeed, the variability in these perturbations was highly similar to that seen in the baseline (random) condition (Figure~\ref{fig:5}\textbf{a}) suggesting our parameter selections were sufficiently robust. Along with these parameter perturbations, we also considered two alternate heuristics: \textit{hot} only or \textit{cold} only. Both of these approaches prevented the strategist from developing any significant influence over the stumbler, leading to no improvement in training performance (Figure~\ref{fig:5}\textbf{b}) compared to the stumbler alone. This confirms that the improved performance of the full network is driven by a search for the optimal heuristic, as opposed to any arbitrary heuristic.

\begin{figure}[bt]
\includegraphics[width=0.33\linewidth]{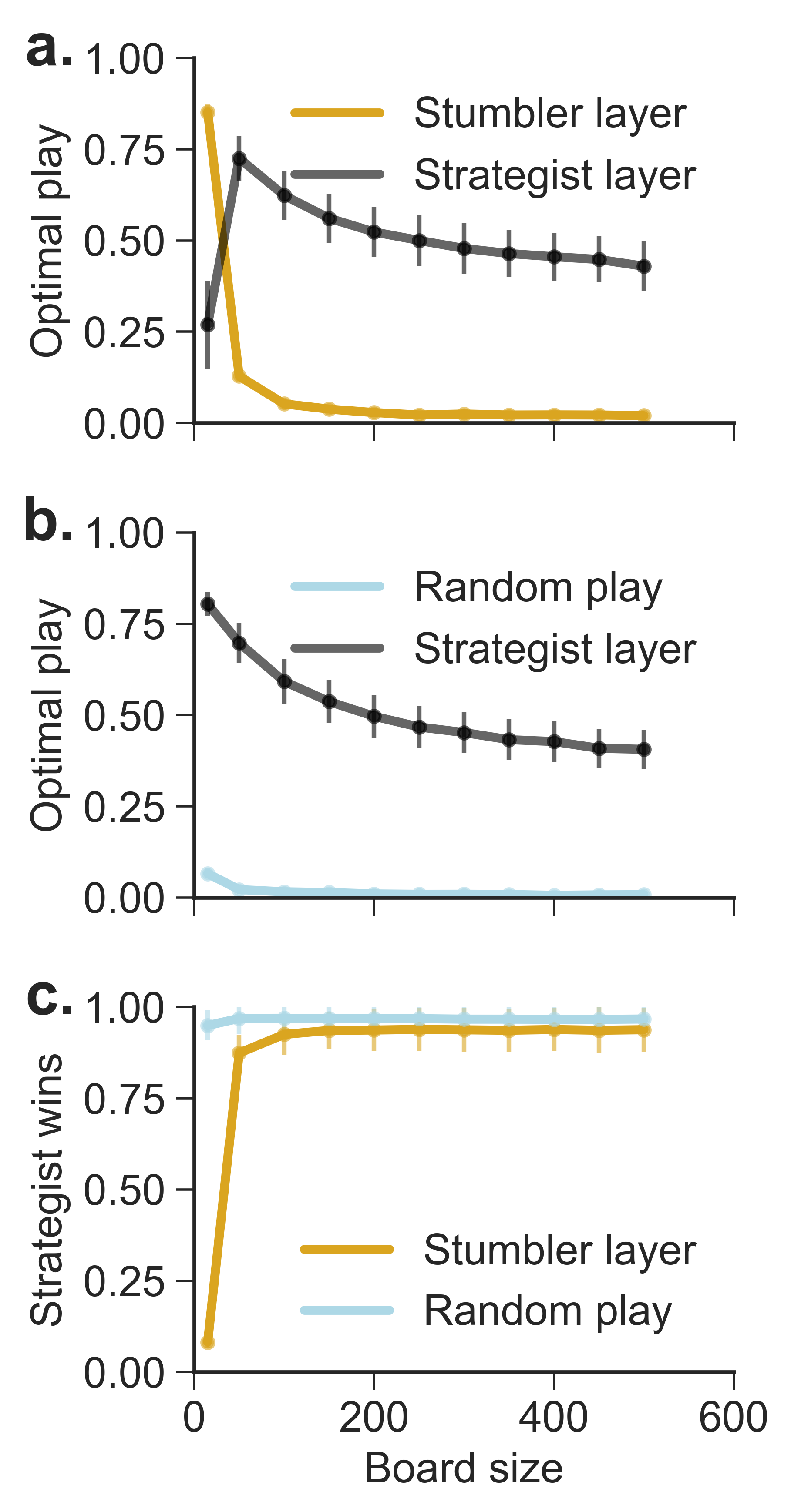} 
\centering
\caption{Transfer performance on new larger game boards, with frozen layers. \textbf{a.} Optimal play observed when the stumbler (goldenrod) and strategist (black) layers complete in several (1000) games of Wythoff's. The x-axis denotes increases in board size. Prior to this experiment, neither layer has experienced a board larger than 15x15 (for the stumbler) or 50x50 (for the strategist. \textbf{b.} The strategist versus a random agent (blue) who's choices were sampled uniformly from the available actions at each position, and could not learn \textbf{c.} Fraction of the games won by the strategist versus the stumbler (goldenrod) or the random player (blue). At each board size 1000 games were played, each begun with a unique random seed.
}
\label{fig:6}
\end{figure}

%\subsection{Game board transfer} 
So far we have shown that adding the strategist drives near optimal performance on never before seen game boards with no additional training. We next wanted to show how the different parts of the network contribute to the generalization across expanded task spaces (i.e., board sizes). For this we systematically increased the board size from size that the stumbler was originally trained on (15x15) up to 500x500 (in 50 unit increments), and evaluated the performance of the different parts of the network independently. As one would expect from a tabular representation, as soon as the board size increased beyond its direct experience, optimal play for the stumbler plummeted (Figure \ref{fig:6}\textbf{a}), quickly matching the performance of a random choice agent (i.e., an agent that simply guesses; Figure \ref{fig:6}\textbf{b}). However on the 15x15 board the stumbler model, with full access to the $Q(s,a)$, outperforms the strategist, which can be seen by comparing the first points in both curves in Figure \ref{fig:6}\textbf{a} and in Figure \ref{fig:6}\textbf{c}. The strategist on the other hand maintains a high level of performance over the entire range of novel game boards showing it is necessary for generalization we observe. To compare performance between layers, we had to temporarily lift the structure that only the stumbler can interact with the board. 

%\subsection{Rule transfer}
Up until this point we have studied Wythoff's game. Next considered how the strategy learned in Wythoff's game could transfer to different grid-world games. To do this we keep the game boards the same as our initial experiments (Figures \ref{fig:4}-\ref{fig:5}) but alter the rules of play in accordance with two other impartial games: Euclid's game and Nim (see \textit{Methods}).

\begin{figure}[bt]
\includegraphics[width=0.4\linewidth]{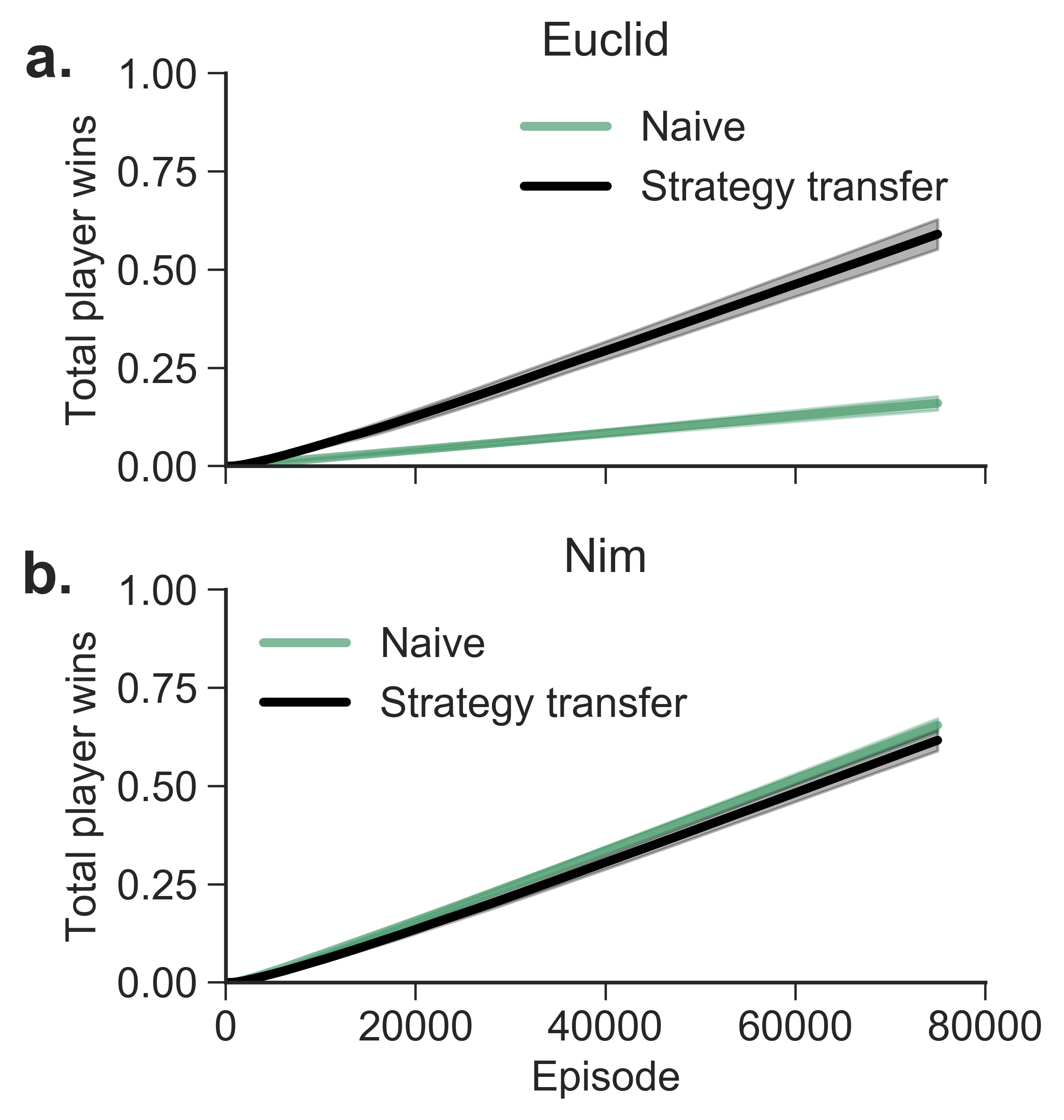} 
\centering
\caption{Rule transfer: learning with a pre-trained strategist. \textbf{a.} Player wins in Euclid's game, with and without strategy transfer. \textbf{b.} Player wins in Nim. In both models only the player could benefit from a strategist. As with all previous simulations, the opponent here was limited to traditional \textit{Q}-learning.}
\label{fig:7}
\end{figure}

Having changed the rules of the game, we studied how a transferring a Wythoff-trained strategy layer, with re-initialized stumbler, impacted learning (Figure \ref{fig:7}). To measure only the benefits of transfer we quantified the ratio of player-to-opponent wins with and without the pre-trained layer. Player models without the pre-trained layer also featured a functioning strategist, though of course it began with a random initial configuration. In all cases and layers learning was ``on''. This means that the pre-trained layer could in princple change and relearn if needed. 

In Euclid's game, whose optimal play is a restricted subset of that of Wythoff's, making it more difficult to learn, strategy transfer not only leads to better performance but actually accelerates learning over a naive stumbler (Figure \ref{fig:7}\textbf{a.}). In contrast, in the game of Nim, strategy transfer does not improve performance, as would be expected given the different optimal strategies, however it also does not significantly hinder learning either. This may be due to the fact that the influence algorithm we employ (Algorithm \ref{algo:strategist-stumbler}) quickly limits the range of bad strategies, leaving the network able to re-train itself quickly in this context.

\begin{figure}[bt]
\includegraphics[width=0.5\linewidth]{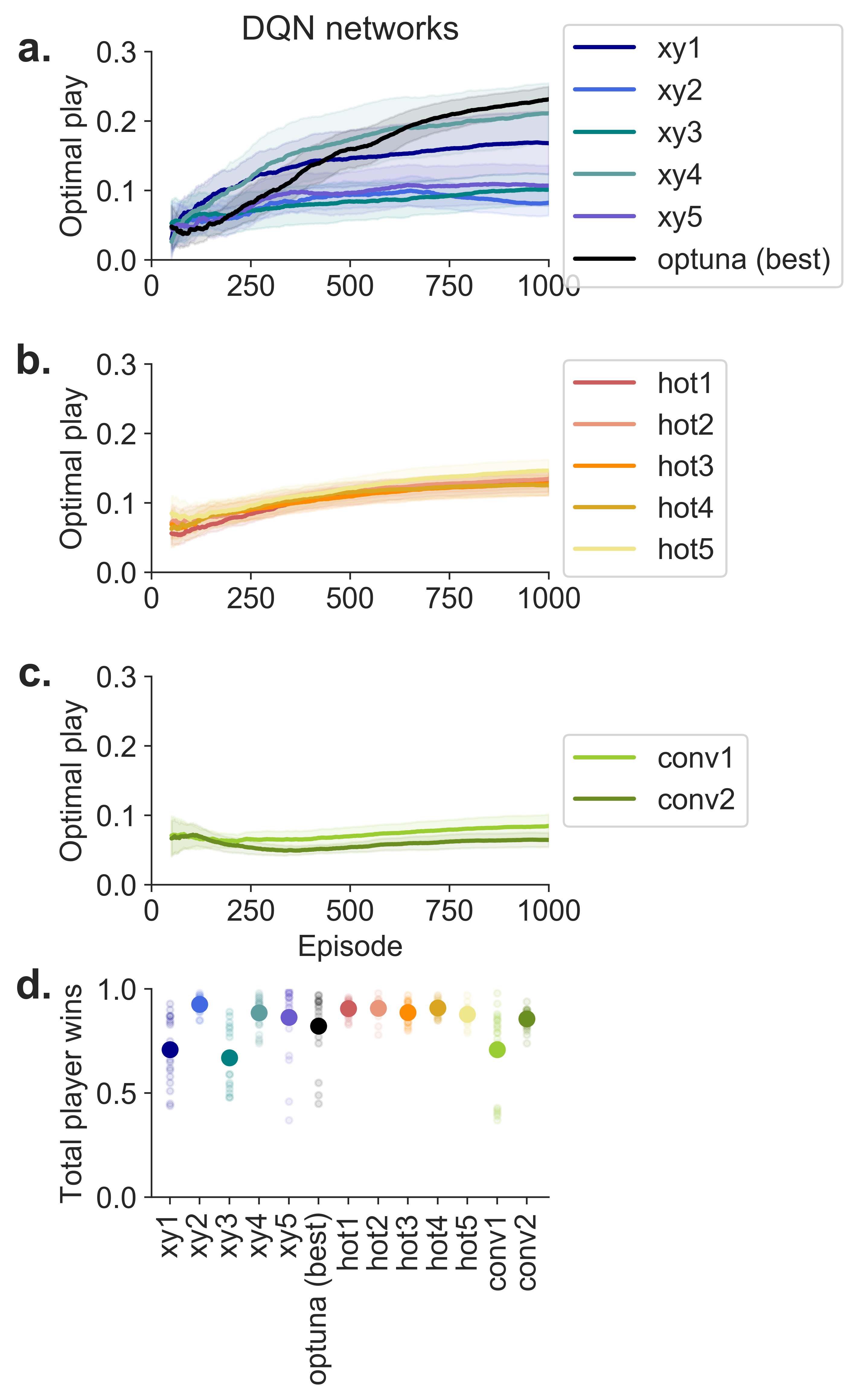} 
\centering
\caption{DQN performance. We tried three styles of network architecture. The first, \textit{xy}, was a MLP whose input is a vector of game position and action, represented as a set of Cartesian coordinates. This is akin to our Strategist model. It's output was a scalar, representing a Q-value estimate. The second, \textit{hot}, was also an MLP, though this one used a one-hot representation. It's input was a $N x N$ vector, that was zeros except for a 1 placed in the current board position. The output was a $N x N$ vector, representing the value of all possible moves at that position. (Illegal moves were masked.) The third, \textit{conv}, had the same input/output representation as \textit{hot}, but featured convolutional layers whose ability to learn local features could, in principle, allow the network to generalize to new board sizes and rules. 
\textbf{a}-\textbf{c}. Optimal play in trained DQNs.
\textbf{d}. Total fraction of wins when trained DQNs played a random opponent ($N$=100). Detailed descriptions of each model can be found in Table~\ref{table:2} and the \textit{Methods}.
} 
\label{fig:9}
\end{figure}

Finally, as a control experiment we trained a series of DQN networks \cite{vanHasselt2015} on Wythoff's game (Figure~\ref{fig:9}). These kinds of models have been described in detail elsewhere \cite{Mnih2013,Mnih2015}. In brief, DQN is the name typically given to a deep neural network who's loss function is based on the Q-learning algorithm, which we've described above. The deep network tries to learn a latent state representation that can interpolate or generalize between board states. It might also be able to learn to generalize optimal play in way that compares well to our approach.

\begin{figure}[bt]
\includegraphics[width=0.8\linewidth]{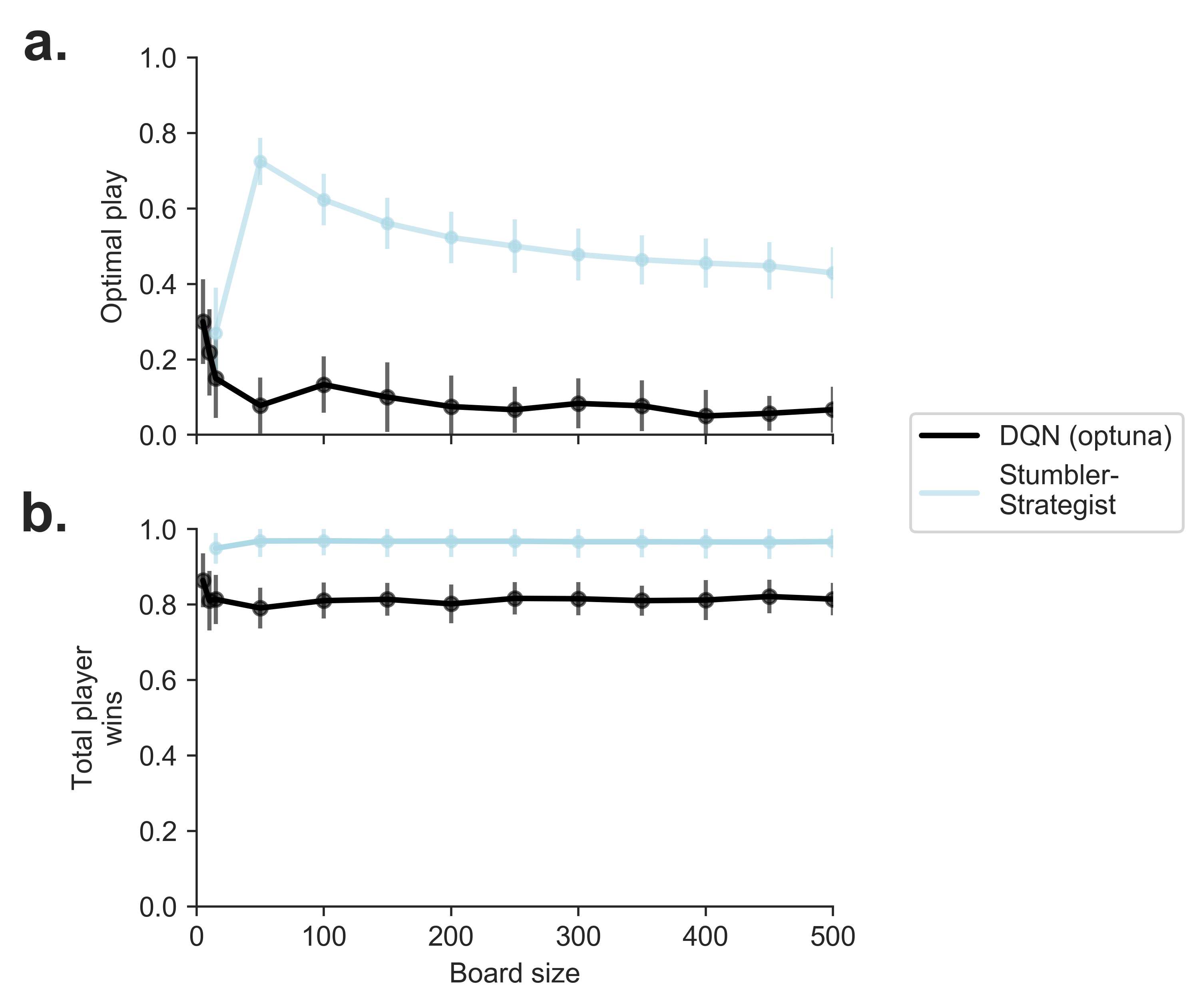}
\centering
\caption{A comparison to deep learning.
\textbf{a}. Optimal play by the best DQN network (black) compared our strategist (teal) on new larger game boards, in Wythoff's ($N=100$). Prior to this experiment both models were trained on a 15x15 board, exclusively.
\textbf{b}. Player wins in versus a random player. The DQN is in black and the strategist is in teal.
Error bars represent SEM.} 
\label{fig:10}
\end{figure}

To try and design a successful DQN model, we tried thirteen network architectures. These included both convolutional (\textit{conv}) networks \cite{Mnih2015} and two styles of MLP, \textit{hot} and \textit{xy} (see, \textit{Methods}). None of these was able to learn to play Wythoffs' optimally (Figure~\ref{fig:9}\textbf{a}-\textbf{c}). Though all networks did learn aspects of the game, and could consistently defeat a random player (Figure~\ref{fig:9}\textbf{d}-\textbf{f}). The best DQN network, when tested on new boards, defeated a random opponent 80-90\% of the time (Figure~\ref{fig:10}\textbf{b}). However its already suboptimal play dropped to near zero (Figure~\ref{fig:10}\textbf{a}). This failure is consistent with similar attempts at using DQNs in other gridworld games \cite{Leike2017}. 

% ---------------------------------------------------------------------------------
% ---------------------------------------------------------------------------------
% ---------------------------------------------------------------------------------
\section{Discussion}
The human prefrontal cortico-basal ganglia pathways are hierarchically organized such that contextual, strategic, and value-based decisions are represented in more anterior regions, while basic action selection and motor control decisions are represented in posterior prefrontal regions, near the motor cortex \cite{badre2009, frank2011mechanisms, Badre2012}. Here we took inspiration \cite{Hassabis2017} from these pathways, as well as their role in imaginative play, to innovate on traditional RL networks by adding a ``strategist'' agent that imagines play in new and more challenging environments and guides a simple RL agent as it learns. The strategist only succeeds in these new, harder, environments because it is learns to predict a simple, but useful, heuristic: every state is good (i.e., hot) or bad (i.e., cold). The combined strategist-stumbler network was able to generalize its learned strategy to changes in action space (i.e., larger game boards) and to different environments where the ideal strategy follows the same heuristic (i.e., different impartial games). Most importantly, however, we could query the strategist to explicitly recover the learned heuristic.

Heuristics offer an advantage for strategic learning because they map complex contingencies to simple rules \cite{Parpart2018}. Heuristics are widely used in human cognition \cite{tversky1974judgment}, having been tuned by evolution and experience to support memory, and transfer knowledge \cite{Gigerenzer2014}. A heuristic is easier to map between environmental states, as it is not dependent on a specific set of complex actions. Intuitively, heuristics also simplify imagining never before experienced outcomes. Though not put to use here, heuristics also ease the transfer of knowledge between agents. For example, it is intuitively easier to transfer knowledge between a student student and teacher, or to engage in cooperative inference, when using simple (appropriate) heuristics rather complex action-value tensors. In this way, exploring how humans and other animals learn heuristics, including the underlying neural substrates, offers unique advantages to building artificial agents that can effectively learn optimal strategies in more complex environments.

Moving from simple games, such as Wythoff's, to open-ended strategy games like Go or chess, and to even more complex visual environments like classic Atari games, will require two further innovations. Gridworld games share common coordinates. It is therefore simple to move from smaller boards to larger and more challenging boards by just mapping between common coordinates. Using a stumbler-strategist network in visually complex environments, like classic Atari games, requires solving this projection problem, which is nontrivial. There is room for optimism though. The response in higher-level layers in DQNs, as well as the inner layers of variational autoencoders, track the perceptual similarity learned images \cite{Zhan2016}. Using these a strategists layer could observe not only values, but also the critical perceptual relationships. 

A critical innovation of the strategist-stumbler model is in its ability to identify the right heuristic. While there seems to be a clear advantages to using heuristics in general, here we studied only a single example (good/bad or hot/cold). We would not expect our choice heuristic to always apply. There exists however a substantial literature in game theory representing a large pool of possible heuristics and strategies \cite{Gigerenzer2014,Parpart2017,Hart2005,Rieskamp2006}. While these will certainly not describe optimal performance in all situations, there are perhaps consistent moments to be found in complex games where simple strategies, and matched counterfactual simulations, will allow for information to be efficiently transferred between environments.  We believe using explicit and human inspired heuristics as ``bottlenecks'' to conditionally simplify complex problems has a general role to play. It can both enhance learning rates, and aide transfer.

Like the problem of reward shaping, putting heuristics to work in complex and more open-ended games (like those available in the OpenAI Gym ecosystem) is not simple. Specifically, there are probably parts of complex games that can benefit from a heuristic bottleneck. But of course there other parts where it won’t help, and might make things worse. Learning to automatically find both points is not simple and there is no reason to bother if the basic approach does’t work in simpler cases like Wythoff's first. %We used a neuro-inspired \cite{Hassabis2017} approach to show how the organizational principles of the human prefrontal cortico-basal ganglia pathways can be leveraged to build artificial agents that learn optimal strategies. 
The success of the strategist-stumbler network at learning the optimal heuristic and generalizing across environments provides a strong proof of concept. Heuristics help both learning and transfer in artificial agents.

\section*{acknowledgements}
The views and conclusions contained in this document are those of the authors and should not be interpreted as representing the official policies, either expressed or implied, of the Army Research Laboratory or the U.S. government.

% Moved from Introduction
%In a previous work \cite{Raghu2017} studied Erdos-Selfridge-Spencer games, which also have an optimal strategy. The goals of our two papers are different. Like \textit{Raghu et al}, we focus on studying transfer compared to an optimal player,  move-by-move. Unlike \textit{Raghu et al}, we try and transfer learning to new environments, rather than focusing only on transferring between changes in opponent strategy. 

% \section*{conflict of interest}
% You may be asked to provide a conflict of interest statement during the submission process. Please check the journal's author guidelines for details on what to include in this section. Please ensure you liaise with all co-authors to confirm agreement with the final statement.

% \printendnotes

% Submissions are not required to reflect the precise reference formatting of the journal (use of italics, bold etc.), however it is important that all key elements of each reference are included.
\bibliography{ss_ejp_library,tim_library}

% \begin{biography}[example-image-1x1]{A.~One}
% Please check with the journal's author guidelines whether author biographies are required. They are usually only included for review-type articles, and typically require photos and brief biographies (up to 75 words) for each author.
% \bigskip
% \bigskip
% \end{biography}

% \graphicalabstract{example-image-1x1}{Please check the journal's author guildines for whether a graphical abstract, key points, new findings, or other items are required for display in the Table of Contents.}

\end{document}